# Kinematics and Workspace Analysis of a Three-Axis Parallel Manipulator: the Orthoglide

A. Pashkevich[1], D. Chablat[2*], P. Wenger[2]

[1]*Robotic Laboratory, Belarusian State University of Informatics and Radioelectronics*
*6 P. Brovka St., Minsk, 220027, Belarus*
*E-mail: pap@ bsuir.unibel.by*

[2]*Institut de Recherche en Communications et Cybernétique de Nantes*
*1 rue de la Noë, BP 92 101, 44321 Nantes, France*
*E-mails: Damien.Chablat@irccyn.ec-nantes.fr, Philippe.Wenger@irccyn.ec-nantes.fr*

**ABSTRACT**

The paper addresses kinematic and geometrical aspects of the Orthoglide, a three-DOF parallel mechanism. This machine consists of three fixed linear joints, which are mounted orthogonally, three identical legs and a mobile platform, which moves in the Cartesian *x-y-z* space with fixed orientation. New solutions to solve inverse/direct kinematics are proposed and we perform a detailed workspace and singularity analysis, taking into account specific joint limit constraints.

KEYWORDS: Parallel manipulators; Workspace isotropy; Inverse and direct kinematics; Singularity.

## 1. INTRODUCTION

For two decades, parallel manipulators attract the attention of more and more researchers who consider them as valuable alternative design for robotic mechanisms[1-3]. As stated by a number of authors[4], conventional serial kinematic machines have already reached their dynamic performance limits, which are bounded by high stiffness of the machine components required to support sequential joints, links and actuators. Thus, while having good operating characteristics (large workspace, high flexibility and manoeuvrability), serial manipulators have disadvantages of low precision, low stiffness and low power. Also, they are generally operated at low speed to avoid excessive vibration and deflection.

---
[*]Corresponding author

Conversely, parallel kinematic machines offer essential advantages over their serial counterparts (lower moving masses, higher rigidity and payload-to-weight ratio, higher natural frequencies, better accuracy, simpler modular mechanical construction, possibility to locate actuators on the fixed base) that obviously should lead to higher dynamic capabilities. However, most existing parallel manipulators have limited and complicated workspace with singularities, and highly non-isotropic input/output relations[5]. Hence, the performances may significantly vary over the workspace and depend on the direction of the motion, which is a serious disadvantage for machining applications.

Research in the field of parallel manipulators began with the Stewart-platform used in flight simulators[6]. Many such structures have been investigated since then, which are composed of six linearly actuated legs with different combinations of link-to-platform connections[7]. In recent years, several new kinematic structures have been proposed that possess higher isotropy. In particular, a 3-dof translational mechanism with gliding foot points was found in three separate works to be fully isotropic throughout the Cartesian workspace[8-9-10]. It consists of a mobile platform, which is connected to three orthogonal linear drives through three identical planar 3-revolute jointed serial chains. Although this manipulator behaves like a conventional Cartesian machine, bulky legs are required to assure stiffness because these legs are subject to bending.

In this paper, the Orthoglide manipulator proposed by Wenger and Chablat[11-13] is studied. As follows from previous research, this manipulator has good kinetostatic performances and some technological advantages, such as (i) symmetrical design consisting of similar 1-d.o.f. joints; (ii) regular workspace shape properties with bounded velocity amplification factor; and (iii) low inertia effects. This paper analyses the kinematics and the workspace of the Orthoglide. Section 2 describes the Orthoglide geometry. Section 3 proposes new solutions for its inverse and direct kinematics. Sections 4,5 present a detailed analysis of the workspace and jointspace respectively. Section 6 contains exhaustive singularity study. And, finally, Section 7 summarises the main contributions of the paper.

## 2. MANIPULATOR GEOMETRY

The kinematic architecture of the Orthoglide is shown in Fig. 1. It consists of three identical kinematic chains that are formally described as *PRP$_a$R*, where *P*, *R* and *P$_a$* denote the prismatic, revolute, and parallelogram joints respectively. The mechanism input is made up by three actuated orthogonal prismatic joints. The output body (with a tool mounting flange) is connected to the prismatic joints through a set of three kinematic chains. Inside each chain, one parallelogram is used and oriented in a manner that the output body is restricted to translational movements only. The arrangement of the joints in the PRPaR chains has been defined to eliminate any constraint singularity[12] in the Cartesian workspace.

To get the Orthoglide kinematic equations, let us locate the reference frame at the intersection of the prismatic joint axes and align the coordinate axis with them (Fig. 2), following the "right-hand" rule. Let us also denote the input vector of the prismatic joints variables as $\boldsymbol{\rho} = (\rho_x, \rho_y, \rho_z)$ and the output position vector of the tool centre point as $\mathbf{p} = (p_x, p_y, p_z)$. Taking into account obvious properties of the parallelograms, the Orthoglide geometrical model can be presented in a simplified form, which consists of three bar links connected by spherical joints to the tool centre point at one side and to the corresponding prismatic joints at another side. Using this notation, the kinematic equations of the Orthoglide can be written as follows

$$
\begin{aligned}
(p_x - \rho_x)^2 + p_y^2 + p_z^2 &= L^2 \\
p_x^2 + (p_y - \rho_y)^2 + p_z^2 &= L^2 \\
p_x^2 + p_y^2 + (p_z - \rho_z)^2 &= L^2
\end{aligned}
\quad (1)
$$

where *L* is the length of the parallelogram principal links and the "zero" position $\mathbf{p}_0 = (0, 0, 0)$ corresponds to the joints variables $\boldsymbol{\rho}_0 = (L, L, L)$, see Fig. 3a.

It should be stressed that the Orthoglide geometry and relevant manufacturing technology impose the following constraints on the joint variables

$$0 < \rho_x \le 2L \ ; \ 0 < \rho_y \le 2L \ ; \ 0 < \rho_z \le 2L, \quad (2)$$

which essentially influence on the workspace shape. While the upper bound ($\rho \leq 2L$) is implicit and obvious, the lower one ($\rho > 0$) is caused by practical reasons, since safe mechanical design encourages avoiding risk of simultaneous location of prismatic joints in the same point of the Cartesian workspace (here and in the following sections, while referring to symmetrical constraints are subscript omitted, i.e. $\rho \in \{\rho_x, \rho_y, \rho_z\}$).

## 3. ORTHOGLIDE KINEMATICS

### 3.1. Inverse kinematics

For the inverse kinematics, the position of the end-point ($p_x$, $p_y$, $p_z$) is treated as known and the goal is to find the joint variables ($\rho_x$, $\rho_y$, $\rho_z$) that yield the given location of the tool. Since in the general case the inverse kinematics can produce several solutions corresponding to the same tool location, the solutions must be distinguished with respect to the algorithm "branch". For instance, if the aim is to generate a sequence of points to move the tool along an arc, care must be taken to avoid branch switching during motion, which may cause inefficient (or even impossible) manipulator motions. Moreover, leg singularities may occur at which the manipulator loses degrees of freedom and the joint variables become linearly dependent. Hence, the complete investigation of the Orthoglide kinematics must cover all the above-mentioned topics.

From the Orthoglide geometrical model (1), the inverse kinematic equations can be derived in a straightforward way as:

$$\begin{aligned} \rho_x &= p_x + s_x \sqrt{L^2 - p_y^2 - p_z^2} \\ \rho_y &= p_y + s_y \sqrt{L^2 - p_x^2 - p_z^2} \\ \rho_z &= p_z + s_z \sqrt{L^2 - p_x^2 - p_y^2} \end{aligned} \qquad (3)$$

where $s_x$, $s_y$, $s_z$ are the branch (or configuration) indices that are equal to ±1. It is obvious that (3) yields eight different branches of the inverse kinematic algorithm, which will be further referred to as *PPP*,

*MPP…MMM* following the sign of the corresponding index (i.e. the notation *MPP* corresponds to the indices $s_x = -1;\ s_y = +1;\ s_z = +1$ ).

The geometrical meaning of these indices is illustrated by Fig. 2, where $\theta_x,\ \theta_y,\ \theta_z$ are the angles between the bar links and the corresponding prismatic joint axes. It can be proved that $s = 1$ if $\theta \in (90^\circ, 180^\circ)$ and $s = -1$ if $\theta \in (0^\circ, 90^\circ)$. That the branch transition ($\theta = 90^\circ$) corresponds to the serial singularity (where the leg is orthogonal to the relevant translational axis and the input joint motion does not produce the end-point displacement).

It is obvious that if the inverse kinematic solution exists, then the target point ($p_x$, $p_y$, $p_z$) belongs to a volume bounded by the intersection of three cylinders

$$C_L = \left\{ \mathbf{p} \mid p_x^2 + p_y^2 \leq L^2;\ p_x^2 + p_z^2 \leq L^2;\ p_y^2 + p_z^2 \leq L^2 \right\} \tag{4}$$

that guarantees non-negative values under the square roots in (3). However, it is not sufficient, since the lower joint limits (2) impose the following additional constraints

$$\begin{aligned} p_x &> -s_x \sqrt{L^2 - p_y^2 - p_z^2}\ ; \\ p_y &> -s_y \sqrt{L^2 - p_x^2 - p_z^2}\ ; \\ p_z &> -s_z \sqrt{L^2 - p_x^2 - p_y^2}\ , \end{aligned} \tag{5}$$

which reduce a potential solution set. For example, it can be easily computed that for the "zero" workspace point $\mathbf{p}_0 = (0, 0, 0)$, the inverse kinematic equations (3) give eight solutions $\mathbf{\rho} = (\pm L, \pm L, \pm L)$ but only one of them is feasible, as shown in Fig. 3.

To analyse in details the influence of the joint constraints impact, let us start from separate a study of the inequalities (5) and then summarise results for all possible combinations of the three configuration indices. If $s_x = 1$, then consideration of two cases, $p_x > 0$ and $p_x \leq 0$, yields the following workspace set satisfying the constraint $\rho_x > 0$

$$W_L^{+x} = \left\{ \mathbf{p} \in C_L \mid p_x > 0 \right\} \cup \left\{ \mathbf{p} \in C_L \mid p_x \leq 0;\ p_x^2 + p_y^2 + p_z^2 < L^2 \right\}, \tag{6}$$

which consists of two fractions (½ of the cylinder intersection denoted $C_L$ and ½ of the sphere whose geometric center is (0,0,0) and radius is $L$). If $s_x = -1$, then the second case $p_x < 0$ does not give any solution and the joint constraint $\rho_x > 0$ is expressed in the workspace as

$$W_L^{-x} = \{\mathbf{p} \in C_L \mid p_x \geq 0;\ p_x^2 + p_y^2 + p_z^2 > L^2\}. \tag{7}$$

The latter defines a solid bounded by three cylindrical surfaces and the sphere. The remaining constrains $\rho_y > 0$ and $\rho_z > 0$ can be derived similarly, which differ from (6), (7) by subscripts only.

Then, there can be found intersection of the obtained sets for different combinations of the configuration indices. It can be easily proved that the case "PPP" yields

$$W_L^{PPP} = \{\mathbf{p} \in C_L \mid p_x, p_y, p_z > 0\} \cup \{\mathbf{p} \in C_L \mid p_x^2 + p_y^2 + p_z^2 < L^2\} \tag{8}$$

while the remaining cases give

$$W_L^{MPP} = ... = W_L^{MMM} = \{\mathbf{p} \in C_L \mid p_x, p_y, p_z > 0\} \cap \{\mathbf{p} \in C_L \mid p_x^2 + p_y^2 + p_z^2 > L^2\} \tag{9}$$

These conclusions can be illustrated by a 2D example presented in Figs. 4 and 5, which show feasible workspace regions for both separate and simultaneous consideration of the constraints on two joint variables $\rho_x, \rho_y$.

Expressions (8) and (9) can be put in the form

$$W_L^{PPP} = S_L \cup G_L; \quad W_L^{MPP} = ... = W_L^{MMM} = G_L \tag{10}$$

where

$$S_L = \{\mathbf{p} \in C_L \mid p_x^2 + p_y^2 + p_z^2 < L^2\};\quad G_L = \{\mathbf{p} \in C_L \mid p_x, p_y, p_z > 0;\ p_x^2 + p_y^2 + p_z^2 > L^2\};\quad S_L \cap G_L = \varnothing.$$

Therefore, for the considered positive joint limits (2), the existence of the inverse kinematic solutions may be summarised as follows:

- Inside the sphere $S_L$ there exist exactly one inverse kinematic solution *PPP* with positive configuration indices $s_x$, $s_y$, $s_z$;

- Outside the sphere $S_L$, but within the positive part of the cylinder intersection $C_L$, there exist 8 solutions of the inverse kinematics (*PPP*, *MPP*, ... *MMM*) corresponding to all possible combinations of the configuration indices $s_x$, $s_y$, $s_z$.

These conclusions may be illustrated by the following numerical examples related to the "unit" manipulator ($L=1$). If the target point **p**=(-0.5, 0.4, 0.3) is within the sphere $S_L$, then the joint coordinates must be taken from the sets $\rho_x \in \{0.37, -1.37\}$, $\rho_x \in \{1.21, -0.41\}$, $\rho_x \in \{1.07, -0.47\}$, which allow only one positive combination. In contrast, for the target point **p**=(0.7, 0.7, 0.7), which is outside the sphere, the inverse kinematics yields solutions with two positive values $\rho_x, \rho_y, \rho_z \in \{0.84, 0.56\}$ that allow 8 positive combinations of the joint variables. An interesting feature is that intermediate cases (with 2 or 4 feasible solutions) are not possible.

### 3.2. Direct kinematics

For the direct kinematics, the values of the joint variables ($\rho_x$, $\rho_y$, $\rho_z$) are known and the goal is to find the tool centre point location ($p_x$, $p_y$, $p_z$) that corresponds to the given joint positions. While, in general, the inverse kinematics of parallel mechanisms is straightforward, the direct kinematics is usually very complex. The Orthoglide has the advantage leave an analytical direct kinematics. Like for the previous section, the solutions must be distinguished with respect to the algorithm "branch" that should be also defined both geometrically and algebraically, via a configuration index.

To solve the system (1) for $p_x$, $p_y$, $p_z$, first, let us derive linear relations between the unknowns. By subtracting three possible pairs of the equations (1), we leave

$$2\rho_x p_x - 2\rho_y p_y = \rho_x^2 - \rho_y^2$$
$$2\rho_x p_x - 2\rho_z p_z = \rho_x^2 - \rho_z^2 \qquad (11)$$
$$2\rho_y p_y - 2\rho_z p_z = \rho_y^2 - \rho_z^2$$

As follows from these expressions, the relation between $p_x$, $p_y$, $p_z$ may be presented as

$$p_x = \frac{\rho_x}{2} + \frac{t}{\rho_x}; \quad p_y = \frac{\rho_y}{2} + \frac{t}{\rho_y}; \quad p_z = \frac{\rho_z}{2} + \frac{t}{\rho_z}, \qquad (12)$$

where *t* is an auxiliary scalar parameter. From a geometrical point of view, the expression (12) defines the set of equidistant points for the prismatic joint centres (Fig. 6). Also, it can be easily proved that the full set of equidistant points is the line perpendicular to $\Pi$ and passing through ($\rho_x, \rho_y, \rho_z$)/2, where.

$$\Pi = \left\{ \mathbf{p} \mid \frac{p_x}{\rho_x} + \frac{p_y}{\rho_y} + \frac{p_z}{\rho_z} = 1 \right\} \qquad (13)$$

After substituting (12) into any of the equations (1), the direct kinematic problem is reduced to the solution of a quadratic equation in the auxiliary variable *t*,

$$At^2 + Bt + C = 0, \qquad (14)$$

where $A = (\rho_x \rho_y)^2 + (\rho_x \rho_z)^2 + (\rho_y \rho_z)^2$; $B = (\rho_x \rho_y \rho_z)^2$; $C = (\rho_x^2/4 + \rho_y^2/4 + \rho_z^2/4 - L^2)(\rho_x \rho_y \rho_z)^2$.

The quadratic formula yields two solutions

$$t = \frac{-B + m\sqrt{B^2 - 4AC}}{2A}; \quad m = \pm 1 \qquad (15)$$

that geometrically correspond to different locations of the target point *P* (see Fig. 6) with respect to the plane passing through the prismatic joint centres (it should be noted that the intersection point of the plane and the set of equidistant point corresponds to $t_0 = -B/(2A)$ ). Hence, the Orthoglide direct kinematics is solved analytically, via the quadratic formula (14) for the auxiliary variable *t* and its substitution into expressions (12).

The direct kinematic solution exists if and only if the joint variables satisfy the inequality $B^2 \geq 4AC$, which defines a closed region in the joint variable space

$$\Re_L = \left\{ \boldsymbol{\rho} \mid \left( \rho_x^2 + \rho_y^2 + \rho_z^2 - 4L^2 \right) \left( \rho_x^{-2} + \rho_y^{-2} + \rho_z^{-2} \right) \leq 1 \right\} \qquad (16)$$

Taking into account the joint limits (2), the feasible joint space may be presented as

$$\Re_L^+ = \left\{ \boldsymbol{\rho} \in \Re_L \mid \rho_x, \rho_y, \rho_z > 0 \right\} \qquad (17)$$

Therefore, for the considered positive joint limits (2), the existence of the direct kinematic solutions may be summarised as follows:

- Inside the region $\Re_L^+$, there exist exactly two direct kinematic solutions, which differ by the target point location relative to the plane $\Pi$ (Fig. 7a);

- On the border of the region $\Re_L^+$ located inside the first octant, there exist a single direct kinematic solution, which corresponds to the "flat" manipulator configuration, where both the target point and prismatic joint centres belong to the plane $\Pi$ (Fig. 7b).

These conclusions may be illustrated by the following numerical examples (for the "unit" manipulator, $L=1$). Since the joint variables $\rho_x=\rho_y=\rho_z=0.3$ are within $\Re_L^+$, then the end-point coordinates are either $p_x=p_y=p_z=-0.46$ or $p_x=p_y=p_z=-0.66$. In contrast, for the joint variables $\rho_x=\rho_y=\rho_z=\sqrt{1.5}$, which are exactly on the surface $\Re_L^+$, the direct kinematics yields a single solution $p_x=p_y=p_z=\sqrt{1/6}$ corresponding to the "flat" configuration (see Fig. 7b).

### 3.3. Configuration indices

As follows from the previous sub-sections, both the inverse and direct kinematics of the Orthoglide may produce several solutions. The problem is how to define numerically the *configuration indices*, which allow choosing among the corresponding algorithm branches.

For the inverse kinematics, when the configuration is defined by the angle between the leg and the corresponding prismatic joint axis, the decision equations for the configuration indices are trivial:

$$s_x = \mathrm{sgn}(\rho_x - p_x); \quad s_y = \mathrm{sgn}(\rho_y - p_y); \quad s_z = \mathrm{sgn}(\rho_z - p_z); \tag{18}$$

Geometrically, $s > 0$ means that $\theta_x, \theta_y, \theta_z \in \left]\dfrac{\pi}{2}, \dfrac{3\pi}{2}\right[$ (see Fig. 2).

For the direct kinematics, the configuration is defined by the end-point location relative to the plane that passes through the prismatic joint centres (see Figs. 6-7). Hence, the decision equation may be derived by analysing the dot-product of the plane normal vector $\left(\rho_x^{-1}, \rho_y^{-1}, \rho_z^{-1}\right)$ and the vector directed along any of the bar links (for instance, $\left(p_x-\rho_x,\ p_y,\ p_z\right)$ for the first link):

$$m = \text{sgn}\left(\frac{p_x}{\rho_x} + \frac{p_y}{\rho_y} + \frac{p_z}{\rho_z} - 1\right) \tag{19}$$

which for the positive joint limits is equivalent to

$$m = \text{sgn}\left(p_x \rho_y \rho_z + \rho_x p_y \rho_z + \rho_x \rho_y p_z - \rho_x \rho_y \rho_z\right) \tag{20}$$

It should be stressed that the feasible solutions for the inverse/direct kinematics, located in the neighbourhood of the "zero" point, have the following configuration indices: $s_x = s_y = s_z = +1$ and $m = -1$.

## 4. WORKSPACE ANALYSIS

The robot workspace is an important criterion in evaluating manipulator performance.

As follows from the equation (10), the Orthoglide workspace $W_L$ is composed of two fractions (Fig. 8): (i) the sphere $S_L$ of radius $L$ and centre point (0, 0, 0), and (ii) the thin non-convex solid $G_L$, which is located in the first octant and bounded by the surfaces of the sphere $S_L$ and the cylinder intersection $C_L$. These surfaces can be generated by applying the following algorithm based on the expressions from Sub-Section 3.2:

---

**Algorithm 1. Orthoglide Workspace (3D Mesh)**

Input:     $\Delta\varphi$, $\Delta\theta$ (grid steps direction of angles $\varphi$, $\theta$)
Output:    $X, Y, Z$ (2D arrays of 3D Face nodes)
**for** $\varphi = 0$ **to** $2\pi$ **step** $\Delta\varphi$
    **for** $\theta = -\pi/2$ **to** $\pi/2$ **step** $\Delta\theta$
        $e_x = \cos\varphi \cos\theta$ ; $e_y = \cos\varphi \sin\theta$ ; $e_z = \sin\varphi$
        **if** ($e_x < 0$) **or** ($e_y < 0$) **or** ($e_z < 0$)
            $k = 1$ ;
        **else**
            $k = \max\left\{\sqrt{e_x^2 + e_y^2};\ \sqrt{e_x^2 + e_z^2};\ \sqrt{e_y^2 + e_z^2}\right\}$
        **end if**
        $X(\varphi, \theta) = e_x L/k$ ; $Y(\varphi, \theta) = e_y L/k$ ; $Z(\varphi, \theta) = e_z L/k$ ;
    **next** $\theta$
**next** $\varphi$

---

where ($e_x$, $e_y$, $e_z$) are the components of a unit direction vector, which are expressed via two angles $\varphi, \theta$.

It can be proved that the volume of $C_L$, $S_L$ and $W_L$ is defined by the expressions

$$Vol(C_L) = 8\left(2 - \sqrt{2}\right) L^3 \approx 4.686\, L^3 \tag{21a}$$

$$Vol(G_L) = \left(2 - \sqrt{2} - \frac{\pi}{6}\right) L^3 \approx 0.062\, L^3 \tag{21b}$$

$$Vol(W_L) = \left(2 + \frac{7\pi}{6} - \sqrt{2}\right) L^3 \approx 4.251\, L^3 \tag{21c}$$

As follows from (21), the Orthoglide with the joint limits (2) uses about 53% of the workspace $V_{PPP}$ of its serial counterpart (a Cartesian PPP machine with $2L \times 2L \times 2L$ workspace). Also, the volume of $G_L$ ($0.062\, L^3$) is insignificant in comparison with the volume of the sphere $S_L$ ($4.189\, L^3$), which is equal to 52% of $V_{PPP}$. On the other hand, releasing the lower joint limit ($\rho > 0$) leads to an increases the workspace volume of up to 59% of $V_{PPP}$ only, since the volume of the workspace is, then, equal to $C_L$.

The mutual location of $G_L$ and $S_L$ (and their size ratio) may be also evaluated by the intersection points of the first octant bisector. In particular, for the sphere $S_L$ the bisector intersection point is located at distance $1/\sqrt{3} \approx 0.58$ from the origin, while for the solid $G_L$ the corresponding distance is $1/\sqrt{2} \approx 0.71$ (assuming that $L=1$). Also, $G_L$ touches the sphere by its circular edges, which are located on the borders of the first octant.

Therefore, the result of the workspace analysis may be summarised as follows:

- The Orthoglide workspace $W_L$ is composed of two fractions, the sphere $S_L$ and the thin non-convex solid $G_L$;
- Inside the sphere $S_L$, there exists a single solution of the inverse kinematics, while within the solid $G_L$ there exist 8 such solutions;
- The total volume of the workspace is about $4.25\, L^3$ that comprises roughly 53% of the workspace of the corresponding serial machine with the same joint limits.

## 5. JOINT SPACE ANALYSIS

The properties of the *feasible* jointspace are essential for the Orthoglide control, in order to avoid impossible combinations of the prismatic joint variables $\rho_x$, $\rho_y$, $\rho_z$, which are generated by the control system and are followed by the actuators. For serial manipulators, this problem does not usually exist because the jointspace is bounded by a parallelepiped and mechanical limitations of the joint values may be verified easily and independently. For parallel manipulators, however, one needs to check both (i) separate input coordinates (to satisfy the joint limits), and (ii) their combinations that must be feasible to produce a direct kinematic solution.

As follows from Sub-Section 3.3, the Orthoglide jointspace $\Re_L^+$ is located within the first octant and is bounded by a surface, which corresponds to a single solution of the direct kinematics. Therefore, the jointspace boundary is defined by the relation $B^2 = 4AC$ (see equation (14)), which may be rewritten as

$$\left(\rho_x^2 + \rho_y^2 + \rho_z^2 - 4L^2\right)\left(\rho_x^{-2} + \rho_y^{-2} + \rho_z^{-2}\right) = 1 \qquad (22)$$

and solved for $\rho_x$ assuming that $\rho_y$, $\rho_z$ are known:

$$D\rho_x^4 + DE\rho_x^2 + E = 0 \qquad (23)$$

where $D = \rho_y^{-2} + \rho_z^{-2}$; $E = \rho_y^2 + \rho_z^2 - 4L^2$.

However, this equation is non-symmetrical with respect to $\rho_x, \rho_y, \rho_z$ and, therefore, is not convenient the real-time control. An alternative way to obtain the jointspace boundary, which is more computationally efficient, is based on the conversion from Cartesian to spherical coordinates

$$\rho_x = e_x t; \quad \rho_y = e_y t; \quad \rho_z = e_z t \qquad (24)$$

where $t \geq 0$ is the length of the vector $\boldsymbol{\rho}$, and ($e_x$, $e_y$, $e_z$) are the components of the unit direction vector, which are expressed via two angles $\varphi, \theta$:

$$e_x = \cos\varphi \cos\theta; \quad e_y = \cos\varphi \sin\theta; \quad e_z = \sin\varphi .$$

where $\varphi, \theta \in \,]0, \pi/2]$.

For such a notation, the original equation (22) is transformed into a linear equation for $t^2$

$$(F-1)t^2 = 4L^2 F \,; \quad F = e_x^{-2} + e_y^{-2} + e_z^{-2} \tag{25}$$

with an obvious solution $t = 2L\sqrt{F/(F-1)}$.

So, the boundary surface for the feasible joint space $\mathfrak{R}_L^+$ can be generated applying the following algorithm, where $\varepsilon$ is a small positive number ($0 < \varepsilon \ll \pi/2$):

---

**Algorithm 2. Orthoglide Jointspace (3D Mesh)**

    Input:       $\Delta\varphi$, $\Delta\theta$ (grid steps direction of angles $\varphi$, $\theta$)

    Output:    $\rho_x$, $\rho_y$, $\rho_z$ (2D arrays of 3D Face nodes)

  **for** $\varphi = \varepsilon$ **to** $\pi/2 - \varepsilon$ **step** $\Delta\varphi$

    **for** $\theta = \varepsilon$ **to** $\pi/2 - \varepsilon$ **step** $\Delta\theta$

        $e_x = \cos\varphi \cos\theta$; $e_y = \cos\varphi \sin\theta$; $e_z = \sin\varphi$

        $F = 1/e_x^2 + 1/e_y^2 + 1/e_z^2$; $t = 2L\sqrt{F/(F-1)}$

        $\rho_x(\varphi,\theta) = e_x t$; $\rho_y(\varphi,\theta) = e_y t$; $\rho_z(\varphi,\theta) = e_z t$;

    **next** $\theta$

  **next** $\varphi$

---

The Orthoglide joint space is presented in Fig. 9. As follows from its analyses, the bounding surface is close to the 1/8th of the sphere $S_{2L}$. At the edges, which are exactly quarters of the circles of the radius $2L$, the surface touches the sphere. However, in the middle, the surface is located out of the sphere. In particular, the intersection point of the first octant bisector is located at the distance $\sqrt{3/2} \approx 1.22$ from the coordinate system origin for the jointspace border and at the distance $2/\sqrt{3} \approx 1.15$ for the sphere $S_{2L}$ (assuming $L=1$).

Hence, the result of the jointspace analysis may be summarised as follows:

- The Orthoglide jointspace $\mathfrak{R}_L^+$ is located within the first octant and is bounded by the surface, which covers the sphere of radius $2L$ and is close to it; also, the boundary surface possesses circular edges that touch the sphere at the borders of the first octant;

- Inside the jointspace, there exist two solutions of the direct kinematics, while on its outer boundary there is a single solution only.

## 6. ORTHOGLIDE SINGULARITIES

Singularities of a robotic manipulator are important features that essentially influence its capabilities. From an engineering point of view, they are exposed as unusual robot behaviour, when the manipulator instantaneously loses or gains degrees of freedom, certain directions of motion are unattainable, or non-zero output motions exist while the actuators are locked. Mathematically, a singular configuration may be defined as ill-conditioning (or rank deficiency) of the Jacobian describing the differential mapping from the jointspace to the workspace (or vice versa).

For the Orthoglide, it is more convenient to express analytically the inverse Jacobian, which is derived in a straightforward way, by differentiating (3) with respect to $p_x$, $p_y$, $p_z$ :

$$\mathbf{J}^{-1}(\mathbf{p}) = \begin{bmatrix} 1 & \dfrac{-s_x p_y}{\sqrt{L^2 - p_y^2 - p_z^2}} & \dfrac{-s_x p_z}{\sqrt{L^2 - p_y^2 - p_z^2}} \\ \dfrac{-s_y p_x}{\sqrt{L^2 - p_x^2 - p_z^2}} & 1 & \dfrac{-s_y p_z}{\sqrt{L^2 - p_x^2 - p_z^2}} \\ \dfrac{-s_z p_x}{\sqrt{L^2 - p_x^2 - p_y^2}} & \dfrac{-s_z p_y}{\sqrt{L^2 - p_x^2 - p_y^2}} & 1 \end{bmatrix} \qquad (26)$$

Taking into account the relations between the input and the output variables, the inverse Jacobian can be also expressed as

$$\mathbf{J}^{-1}(\mathbf{p}, \boldsymbol{\rho}) = \begin{bmatrix} 1 & \dfrac{p_y}{p_x - \rho_x} & \dfrac{p_z}{p_x - \rho_x} \\ \dfrac{p_x}{p_y - \rho_y} & 1 & \dfrac{p_z}{p_y - \rho_y} \\ \dfrac{p_x}{p_z - \rho_z} & \dfrac{p_y}{p_z - \rho_z} & 1 \end{bmatrix} \qquad (27)$$

that corresponds to the Orthoglide Jacobian presentation proposed by Chablat and Wenger[12,13]. Using (26), the Jacobian determinant may be expressed as

$$\det(\mathbf{J}^{-1}) = \frac{p_x \rho_y \rho_z + \rho_x p_y \rho_z + \rho_x \rho_y p_z - \rho_x \rho_y \rho_z}{(p_x - \rho_x)(p_y - \rho_y)(p_z - \rho_z)} \qquad (28)$$

To perform the analysis, let us distinguish two cases, $\det(\mathbf{J})=0$ and $\det(\mathbf{J}^{-1})=0$, corresponding to different types of the singularities.

If $\det(\mathbf{J})=0$ (*serial,* or *inverse kinematic singularity*), the mapping from the joint velocity space to the tool velocity space is ill-conditioned. It means that certain directions of motion are unattainable and the manipulator loses at least one degree of freedom. The corresponding relations between the manipulator variables are:

$$(p_x=\rho_x) \quad \text{or} \quad (p_y=\rho_y) \quad \text{or} \quad (p_z=\rho_z) \tag{29}$$

$$p_x\rho_y\rho_z + \rho_x p_y\rho_z + \rho_x\rho_y p_z - \rho_x\rho_y\rho_z \neq 0 \tag{30}$$

Geometrically, this type of singularity corresponds to the orthogonal orientation of the parallelogram links relative to the relevant prismatic joint axes (i.e. $p_x=\rho_x$, $\theta_x=\pi/2$, etc.; see Fig. 2) and the work point $P$ is located on the corresponding surface of the cylinder $C_L$. As follows from the workspace analysis, such points belong to the external border of the thin non-convex solid $G_L$ (Fig. 10, singularity "*a*").

If $\det(\mathbf{J}^{-1})=0$ (*parallel,* i.e. *direct kinematic singularity*), the mapping from the tool velocity space to the joint velocity space is ill-conditioned. It means that the manipulator loses instantaneously its stiffness and certain output motion may exist while the actuators are locked. The corresponding relations between the Orthoglide variables are derived from the numerator of (27):

$$p_x\rho_y\rho_z + \rho_x p_y\rho_z + \rho_x\rho_y p_z - \rho_x\rho_y\rho_z = 0 \tag{31}$$

$$(p_x \neq \rho_x) \quad \text{and} \quad (p_y \neq \rho_y) \quad \text{and} \quad (p_z \neq \rho_z) \tag{32}$$

Let us analyse (31) using the direct kinematic solution derived in sub-section 3.2. If the joint variables are non-zero ($\rho_x,\rho_y,\rho_z \neq 0$), then substitution of (12) into (31) gives the following equation in the auxiliary variable $t$

$$\frac{1}{2}\rho_x\rho_y\rho_z + \left(\frac{\rho_x\rho_y}{\rho_z} + \frac{\rho_x\rho_z}{\rho_y} + \frac{\rho_y\rho_z}{\rho_x}\right)t = 0, \tag{33}$$

with a root

$$t_0 = -\frac{\left(\rho_x \rho_y \rho_z\right)^2}{2\left(\rho_x^2 \rho_y^2 + \rho_x^2 \rho_z^2 + \rho_y^2 \rho_z^2\right)}, \tag{34}$$

which exactly corresponds to $t_0 = -B/(2A)$, i.e. to the single solution case of the direct kinematics (see Eq. (15) ). In the opposite case, when at least one of the joint variables is equal to zero, there exist a number of other singularity points that belong to the sphere $S_L$ boundary $\|\mathbf{p} - \mathbf{p}_0\| = L$.

Geometrically, hence, the direct kinematic singularity may be subdivided into two sub-cases. The first of them corresponds to the "flat" manipulator configuration, when the target point and centres of the prismatic joints belong to the same plane (Fig 10, singularity "*b*"). The second one corresponds to the "bar" configuration, when all three links are parallel to each other (Fig 10, singularity "*c*"), i.e. the links are aligned for the simplified model used in this paper, or "half-bar" posture, when two of three links are aligned.

Using these results, let us investigate in detail singularities, which occur on the sphere $S_L$ boundary, positive 1/8 part of which is located inside of the Orthoglide workspace and the remaining 7/8 bounds the manipulator workspace (see Fig. 10). Since the corresponding Cartesian coordinates satisfy the equation $p_x^2 + p_y^2 + p_z^2 = L$, the inverse kinematics (3) yields eight solutions, which may be expressed as $\rho_x = p_x \pm p_x$; $\rho_x = p_x \pm p_x$; $\rho_z = p_x \pm p_x$. These solutions are summarised in Table 1, which also contains expressions for the inverse Jacobian determinant derived from (28). It should be noted that the solutions $\rho_2 \ldots \rho_8$ are feasible only for the positive workspace octant $p_x, p_y, p_z > 0$ (otherwise, the constraints on the joint variables (2) are violated), while the first ("bar" posture) solution $\rho_1$ is feasible throughout the whole sphere $S_L$ boundary. Besides, as follows from the Table 1, some points of the $S_L$ boundary demonstrate an interesting feature: *simultaneous occurrence of the direct and inverse singularities*. This feature occurs for the quarters of the circles $\mathbf{p} = (0, L\cos\alpha, L\sin\alpha)$, $\mathbf{p} = (L\cos\alpha, L\sin\alpha, 0)$, $\mathbf{p} = (L\cos\alpha, 0, L\sin\alpha)$, $\alpha \in [0, \pi/2]$, where the solutions $\rho_5, \rho_6, \rho_7$ create specific cases of the "flat posture" where one of the links is orthogonal to its joint axis. However, it is worth observing that only 1/8 of the $S_L$ boundary belongs to the workspace (see Fig. 10). It should be stressed that the analytical expression for the singularity points $(p_x, p_y, p_z)$ may be derived directly from (3), (31) and rewritten as

$$2p_x p_y p_z + p_x p_y \sqrt{L^2 - p_x^2 - p_y^2} + p_x p_z \sqrt{L^2 - p_x^2 - p_z^2} + p_y p_z \sqrt{L^2 - p_y^2 - p_z^2} =$$
$$= \sqrt{L^2 - p_x^2 - p_y^2} \cdot \sqrt{L^2 - p_x^2 - p_z^2} \cdot \sqrt{L^2 - p_y^2 - p_z^2} \qquad (35)$$

However, it is very tedious and not convenient for practical applications. For this reason, the singularity surface (Fig. 11) may be generated numerically, using the following algorithm:

---

**Algorithm 3. Orthoglide Singularity Surface (3D Mesh)**

Input:   $\Delta\varphi, \Delta\theta$  (grid steps direction of angles $\varphi, \theta$)

Output:  $X, Y, Z$  (2D arrays of 3D Face nodes)

**for** $\varphi = \varepsilon$ **to** $\pi/2 - \varepsilon$ **step** $\Delta\varphi$

  **for** $\theta = \varepsilon$ **to** $\pi/2 - \varepsilon$ **step** $\Delta\theta$

  $e_x = \cos\varphi \cos\theta$;  $e_y = \cos\varphi \sin\theta$;  $e_z = \sin\varphi$

  $F = 1/e_x^2 + 1/e_y^2 + 1/e_z^2$;  $t = 2L\sqrt{F/(F-1)}$

  $t_0 = -(e_x e_y e_z)^2 / 2(e_x^2 e_y^2 + e_x^2 e_z^2 + e_y^2 e_z^2)$

  $X(\varphi,\theta) = (e_x/2 + t_0/e_x) \cdot t$;  $Y(\varphi,\theta) = (e_y/2 + t_0/e_y) \cdot t$;  $Z(\varphi,\theta) = (e_z/2 + t_0/e_z) \cdot t$

  **next** $\theta$

**next** $\varphi$

---

As follows from the analysis, the "flat" singularity surface is located inside the sphere $S_L$ and, at the edges, which are quarters of the circles of radius $L$, the surface touches the sphere. In particular, the intersection point of the first octant bisector is located at the distance $1/\sqrt{6} \approx 0.41$ from the origin for the singularity surface and at the distance $1/\sqrt{3} \approx 0.58$ for the sphere $S_L$ (assuming that $L=1$). For the "bar" singularity, it is obvious that the corresponding points are located on the sphere $S_L$.

For design purposes, it is also useful to evaluate workspace properties along specific directions or within specific cutting planes. Such plots are presented in Figs. 12, 13. The first of them (Fig. 12) illustrates the location of characteristic points along the bisector line $p_x = p_y = p_z$,

$$\mathbf{p}_1 = -1/\sqrt{3}\,\mathbf{e} \approx -0.58\,\mathbf{e};  \quad \mathbf{p}_2 = 1/\sqrt{6}\,\mathbf{e} \approx 0.41\,\mathbf{e};$$
$$\mathbf{p}_3 = 1/\sqrt{3}\,\mathbf{e} \approx 0.58\,\mathbf{e};  \quad \mathbf{p}_4 = 1/\sqrt{2}\,\mathbf{e} \approx 0.71\,\mathbf{e}. \qquad (36)$$

which are expressed via the unit vector $\mathbf{e} = (1,\ 1,\ 1)$. The second illustration (Fig. 13) shows the evolution of the Jacobian determinant $\det(\mathbf{J})$ while the manipulator is moving along this line. As follows from the analysis, it is prudent to limit the workspace by the parallel singularity surface (see Fig. 11) that extracts a fraction (of about 4.8%) of the sphere $S_L$ (Fig. 14), so the total volume reduces up to $4.07\,L^3$. This conclusion is also confirmed by Fig. 15, which shows the evolution of the inverse condition number[14] of the Jacobian $cond(\mathbf{J})^{-1}$ for the workspace horizontal sections.

Hence, results for the singularity analysis may be summarised as follows:

- The Orthoglide workspace includes two types of singular positions that cause degeneration of the inverse/direct kinematic relations; they may be identified as $\det(\mathbf{J}) = 0$ and $\det(\mathbf{J}^{-1}) = 0$;

- The first type (*serial, or inverse kinematic singularity*), for which $\det(\mathbf{J}) = 0$, corresponds to the orthogonal orientation of the parallelogram links relative to the prismatic joint axes; such positions are located on the external border of the set $G_L$;

- The second type (*parallel, or direct kinematic singularity*), for which $\det(\mathbf{J}^{-1}) = 0$, may be subdivided into 2 sub-cases:

   (i) "*flat*" manipulator configuration, when the target point and centres of the prismatic joints belong to the same plane; such positions are located on the surface, which is located inside the sphere $S_L$ and touches it at the edges;

   (ii) "*bar*" manipulator configuration, when all three links are aligned; such positions are located on the sphere $S_L$.

- For practical applications, the workspace should include only singularity-free points, which belong to the part of the sphere $S_L$ with volume $4.07\,L^3$ that is bounded by the parallel singularity surface in the first octant (see Fig. 14); such workspace reduction from 53% to 51% of $V_{PPP}$ also ensures the unique inverse/direct kinematics with the configuration indices $s_x = s_y = s_z = 1$ and $m = -1$ respectively.

## 7. CONCLUSIONS

This paper focuses on the kinematics and workspace analysis of the Orthoglide, a 3-DOF parallel mechanism with a kinematic behaviour close to the conventional Cartesian machine. In contrast to the previous works, the joint variables are assumed to be subject to the specific manufacturing constraints. It leads to essential revising of the manipulator workspace structure, which becomes non-symmetric and has less inverse kinematics solutions. Also, we have proposed a formal definition of the configuration indices and new simple analytical expressions for the Orthoglide inverse/direct kinematics. We have provided algorithms to compute the Cartesian workspace, the joint space and the singular configuration surface.

It was proved that, for the considered joint limits, the Orthoglide workspace is composed of two fractions, the sphere and a thin non-convex solid in which there are 1 and 8 inverse kinematic solutions, respectively. The total workspace volume comprises about 53% of the corresponding serial machine workspace, where over 52% belongs to the sphere (for comparison, releasing of the joint limits yields to an increase of up to 59% in the workspace volume). It is also showed, that the Orthoglide jointspace is bounded by surface with circular edges, which is more convex than the sphere but is rather close to it. Within the jointspace, there exist exactly two direct kinematic solutions, while on the boundary surface there is only one solution.

The Orthoglide workspace includes two types of singular positions that cause degeneration of the inverse/direct kinematic relations. The first type (*serial, or inverse one*), corresponds to the orthogonal orientation of the parallelogram links relative to the prismatic joint axes; such positions are located on the external border of the thin non-convex solid. The second type (*parallel, or direct one*), may be subdivided into 2 sub-cases: (i) the "*flat*" manipulator configuration, when the end-point and centres of the prismatic joints belong to the same plane; and (ii) the "*bar*" manipulator configuration, when all three links are aligned. For the flat configuration, the workspace points are located inside the sphere, while for the bar configuration they belong to the sphere surface. For real-life applications, only the singularity-free region of the workspace should be used. This volume represents 97.2% of the sphere and is bounded by the "flat" singularity surface in the first octant.

These results can be further used for the optimisation of the Orthoglide parameters, which is the subject of our future work.

**FIGURES**

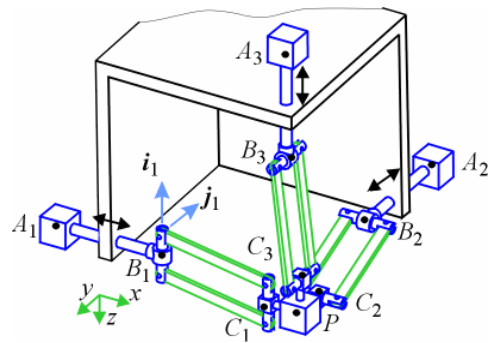

Fig. 1. Orthoglide kinematic architecture.

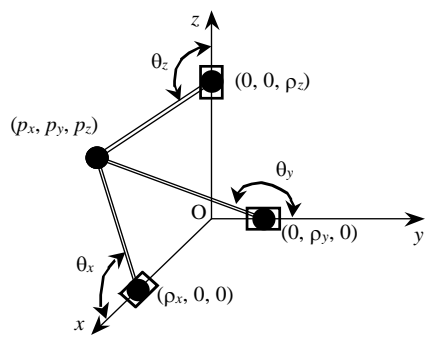

Fig. 2. Orthoglide geometrical model .

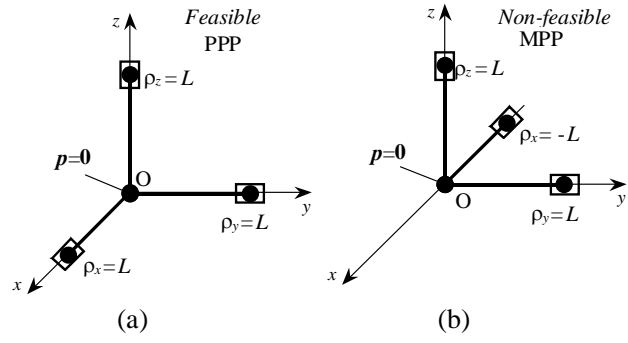

Fig. 3. Feasible and non-feasible "zero" configurations.

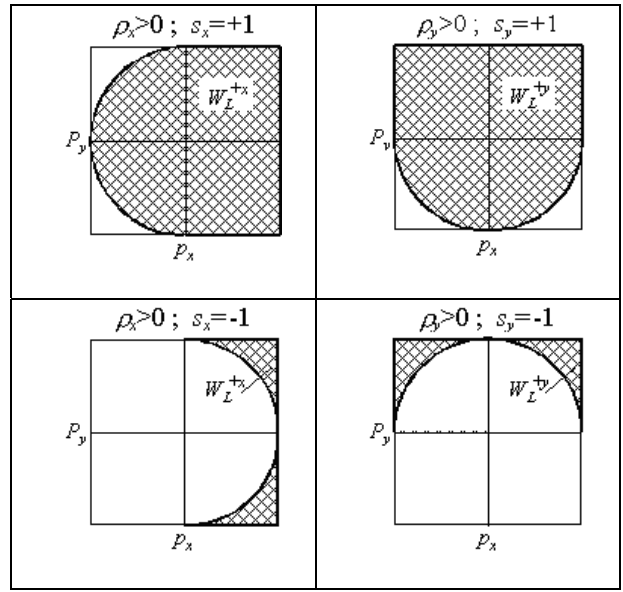

Fig. 4. Feasible regions for separate constraints (2D example)

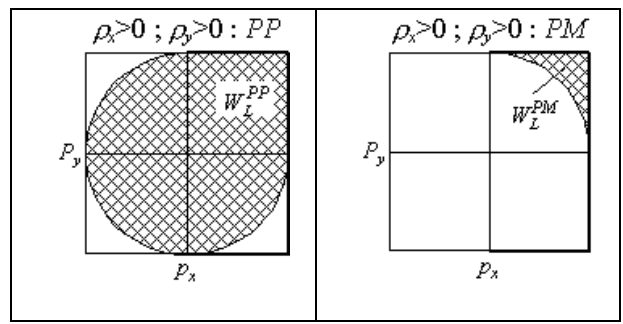

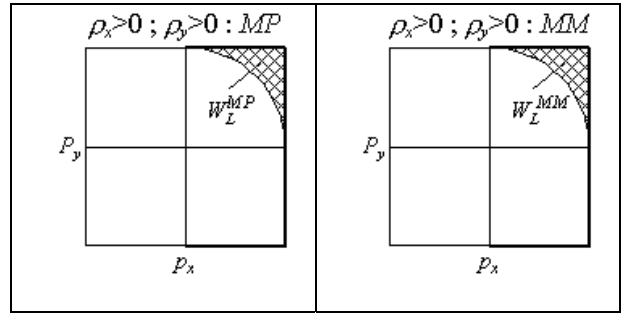

Fig. 5. Feasible regions for simultaneous constraints (2D example)

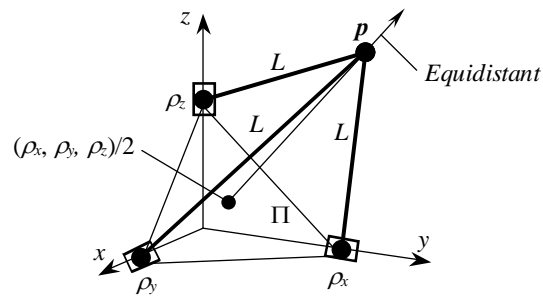

Fig. 6. Geometrical solution of the direct kinematics.

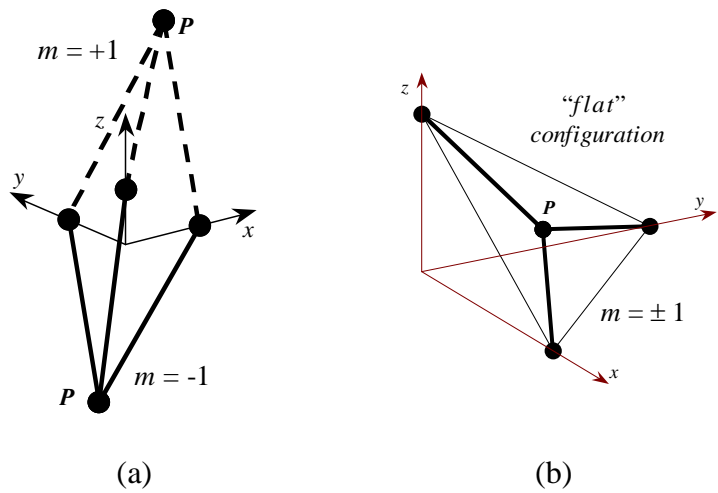

Fig. 7. Double (a) and single (b) solutions of the direct kinematics.

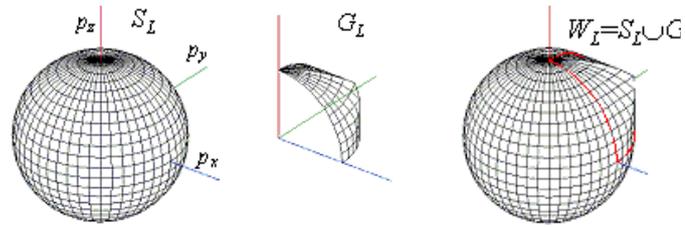

Fig 8. Orthoglide workspace.

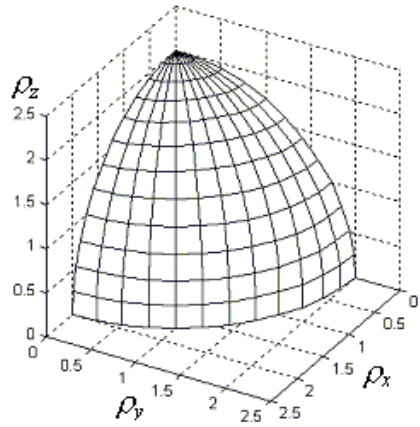

Fig 9. Feasible joint space.

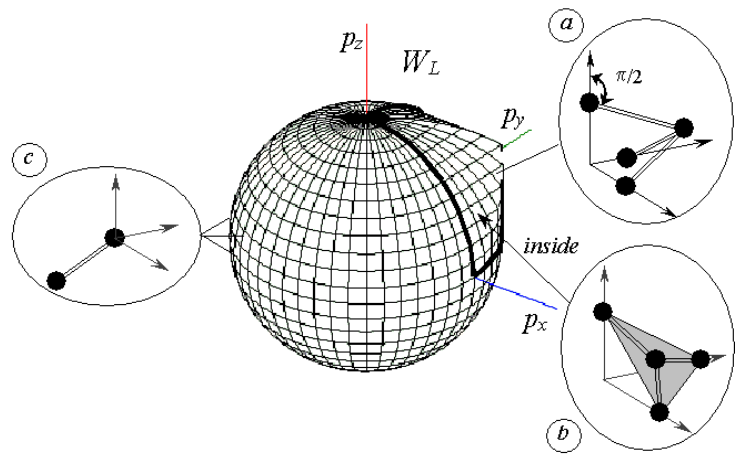

Fig. 10. Geometrical interpretation of the singularities.

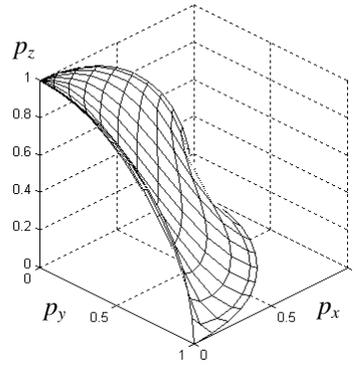

Fig. 11. Parallel ("flat") singularity surface.

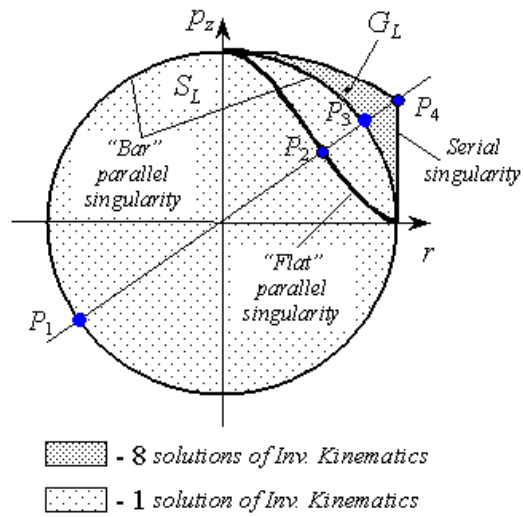

Fig. 12. Cross-section of the workspace ( $p_x = p_y$ ; $r = \sqrt{p_x^2 + p_y^2}$ ).

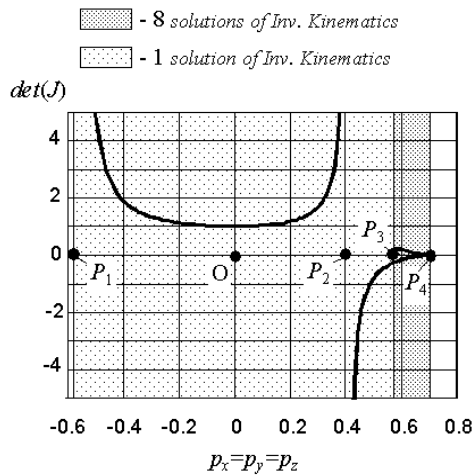

Fig. 13. Evolution of the $\det(\mathbf{J})$ in the bisector direction (for multiple solutions, the determinants differ only by the sign).

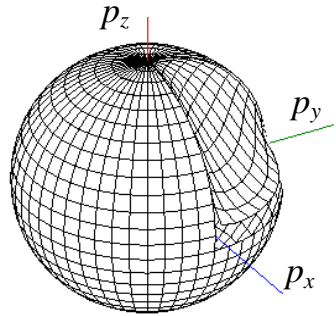

Fig. 14. Orthoglide singularity-free workspace (97.2% of the sphere).

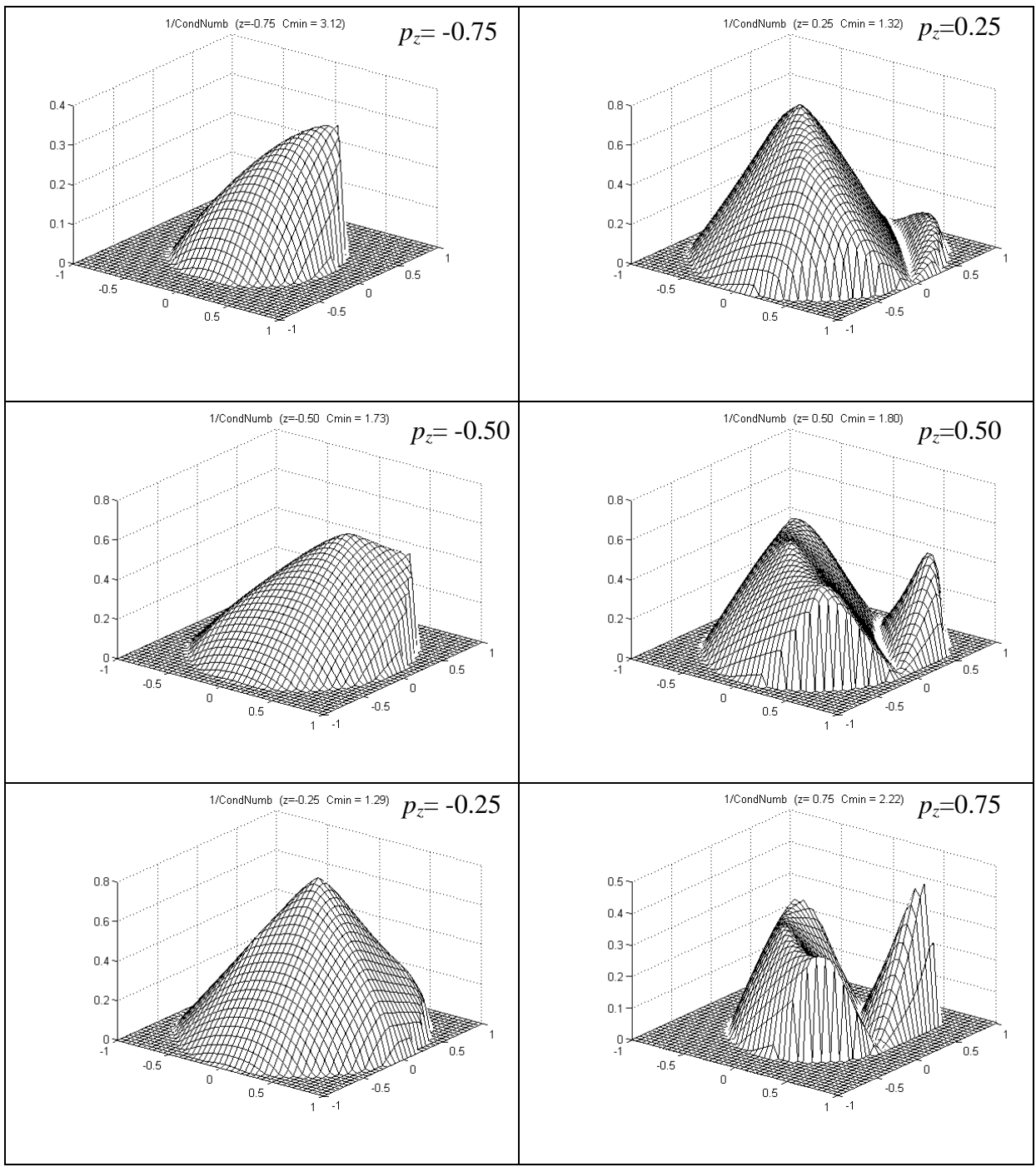

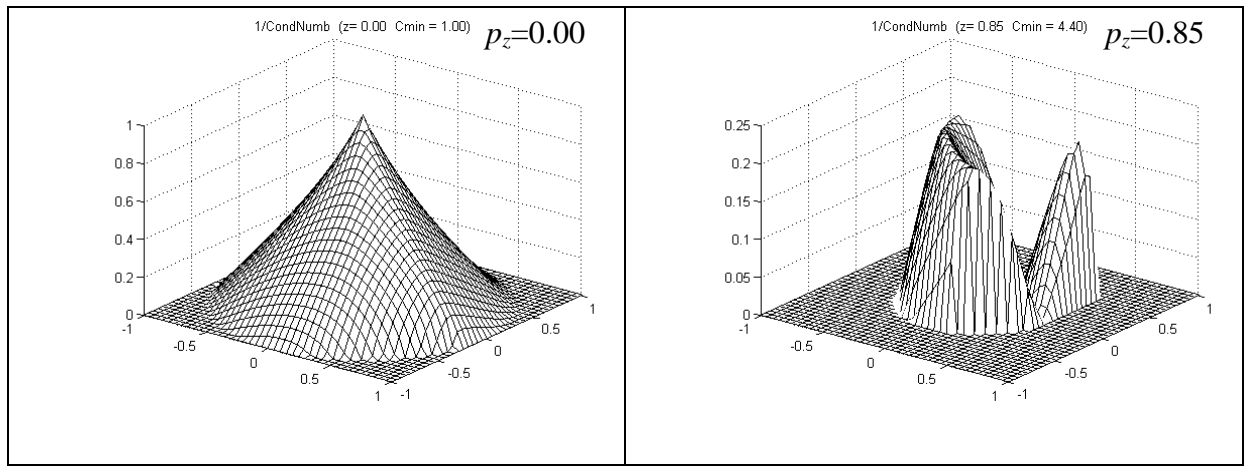

Fig. 15. Evolution of the $cond(\mathbf{J})^{-1}$ for the horizontal sections.

Table 1. Orthoglide postures for the boundary of the sphere $S_L$ boundary.

| Joint coordinates, $\rho$ | Inverse Jacobian $\det(\mathbf{J}^{-1})$ | Posture & singularities |
|---|---|---|
| $\boldsymbol{\rho}_1 = (0, 0, 0)$ | $\dfrac{0}{p_x p_y p_z}$ | "bar" posture (three legs are aligned)<br><br>*direct singularity* (all cases), *inverse singularity*<br><br>(only if $p_x = 0$ or $p_y = 0$ or $p_z = 0$) |
| $\boldsymbol{\rho}_2 = (2p_x, 0, 0)$<br><br>$\boldsymbol{\rho}_3 = (0, 2p_y, 0)$<br><br>$\boldsymbol{\rho}_4 = (0, 0, 2p_z)$ | $\dfrac{0}{-p_x p_y p_z}$ | "half-bar" posture (two legs are aligned)<br><br>*direct singularity* (all cases), *inverse singularity*<br><br>(only if $p_x = 0$ or $p_y = 0$ or $p_z = 0$) |
| $\boldsymbol{\rho}_5 = (2p_x, 2p_y, 0)$<br><br>$\boldsymbol{\rho}_6 = (2p_x, 0, 2p_z)$<br><br>$\boldsymbol{\rho}_7 = (0, 2p_y, 2p_z)$ | $\dfrac{4 p_x p_y p_z}{p_x p_y p_z}$ | *non-singular* general posture if<br><br>$p_x \neq 0$ and $p_y \neq 0$ and $p_z \neq 0$)<br><br>"flat" *singular* posture (both *inverse and direct*) if<br><br>$p_x = 0$ or $p_y = 0$ or $p_z = 0$) |
| $\boldsymbol{\rho}_8 = (2p_x, 2p_y, 2p_y)$ | $\dfrac{4 p_x p_y p_z}{-p_x p_y p_z}$ | *non-singular* general posture if<br><br>$p_x \neq 0$ and $p_y \neq 0$ and $p_z \neq 0$)<br><br>described above singular case,<br><br>if $p_x = 0$ or $p_y = 0$ or $p_z = 0$ |

**FIGURE CAPTIONS**

Fig. 1. Orthoglide kinematic architecture

Fig. 2. Orthoglide geometrical model .

Fig. 3. Feasible and non-feasible "zero" configurations.

Fig. 4. Feasible regions for separate constraints (2D example)

Fig. 5. Feasible regions for simultaneous constraints (2D example)

Fig. 6. Geometrical solution of the direct kinematics .

Fig. 7. Double (a) and single (b) solutions of the direct kinematics.

Fig 8. Orthoglide workspace.

Fig 9. Feasible joint space.

Fig. 10. Geometrical interpretation of the singularities.

Fig. 11. Parallel ("flat") singularity surface.

Fig. 12. Cross-section of the workspace ( $p_x = p_y$ ; $r = \sqrt{p_x^2 + p_y^2}$ ).

Fig. 13. Evolution of the $\det(\mathbf{J})$ in the bisector direction.

Fig. 14. Orthoglide singularity-free workspace (97.2% of the sphere).

Fig. 15. Evolution of the $cond(\mathbf{J})^{-1}$ for the horizontal sections.